\pdfoutput=1

\documentclass[11pt]{article}

\usepackage{EMNLP2022}

\usepackage{times}
\usepackage{latexsym}

\usepackage[T1]{fontenc}

\usepackage[utf8]{inputenc}

\usepackage{microtype}

\usepackage{inconsolata}

\usepackage{amsmath}
\usepackage{times}
\usepackage{framed}
\usepackage{ragged2e}
\usepackage{bm}
\usepackage{soul}
\usepackage{latexsym}
\usepackage{subfigure}
\usepackage{amsthm}
\usepackage{graphicx}
\usepackage{float}
\usepackage{longtable}
\usepackage{booktabs} 
\usepackage{multirow}
\usepackage{multicol}
\usepackage{amssymb}
\usepackage{dashbox}%
\usepackage{multirow}
\usepackage{textcomp}
\usepackage{tcolorbox}
\usepackage{bbding}
\usepackage[ruled, vlined, linesnumbered]{algorithm2e}
\usepackage{mathtools}
\usepackage{adjustbox}
\usepackage[ruled,vlined]{algorithm2e}
\usepackage{color, soul}
\usepackage{array}
\newcolumntype{P}[1]{>{\centering\arraybackslash}p{#1}}
\newcolumntype{M}[1]{>{\centering\arraybackslash}m{#1}}
\usepackage{cleveref}
\crefname{section}{§}{§§}
\Crefname{section}{§}{§§}
\crefname{figure}{Figure}{Figure}
\Crefname{figure}{Figure}{Figure}
\crefname{table}{Table}{Table}
\Crefname{table}{Table}{Table}
\definecolor{forestgreen}{HTML}{228B22}

\newcommand\ourdata{\textsc{Maven-Ere}\xspace}
\newcommand\mycheck{\textcolor{forestgreen}{\Checkmark}\xspace}
\newcommand\myx{\textcolor{red}{\XSolidBrush}\xspace}
\newcommand\RBT{RoBERTa$_{\textsc{Base}}$\xspace}

\title{\ourdata: A Unified Large-scale Dataset for Event Coreference, Temporal, Causal, and Subevent Relation Extraction}

\author{ Xiaozhi~Wang$^{1}\thanks{\quad indicates equal contribution.}$\hspace{0.5em}, Yulin~Chen$^{2*}$, Ning~Ding$^{1}$, Hao~Peng$^{1}$, Zimu~Wang$^5$, Yankai~Lin$^{6,7}\thanks{\quad Partly done while Y.Lin and P.Li were at Tencent.}$\hspace{0.5em}, \\ \textbf{Xu~Han$^{1}$, Lei~Hou$^{1}$, Juanzi~Li$^{1,3}\thanks{\quad Corresponding author: J.Li.}$\hspace{0.5em},} \textbf{Zhiyuan~Liu$^{1}$}, \textbf{Peng~Li$^{4\dag}$, Jie~Zhou$^8$} \\
 $^1$Department of Computer Science and Technology, BNRist;\\
 $^2$Shenzhen International Graduate School;\\
 $^3$THU-Siemens Ltd., China Joint Research Center for Industrial Intelligence and IoT;\\
 $^4$Institute for AI Industry Research (AIR), Tsinghua University, Beijing, China \\
 $^5$Xi'an Jiaotong-Liverpool University, Suzhou, China \\
 $^6$Gaoling School of Artificial Intelligence, Renmin University of China, Beijing, China \\
 $^7$Beijing Key Laboratory of Big Data Management and Analysis Methods, Beijing, China \\
 $^8$Pattern Recognition Center, WeChat AI, Tencent Inc, China \\
 \texttt{\{wangxz20,yl-chen21\}@mails.tsinghua.edu.cn}
 }

\begin{document}
\maketitle
\begin{abstract}
The diverse relationships among real-world events, including coreference, temporal, causal, and subevent relations, are fundamental to understanding natural languages. However, two drawbacks of existing datasets limit event relation extraction (ERE) tasks: (1) Small scale. Due to the annotation complexity, the data scale of existing datasets is limited, which cannot well train and evaluate data-hungry models. (2) Absence of unified annotation. Different types of event relations naturally interact with each other, but existing datasets only cover limited relation types at once, which prevents models from taking full advantage of relation interactions.  To address these issues, we construct a unified large-scale human-annotated ERE dataset \ourdata with improved annotation schemes. It contains $103,193$ event coreference chains, $1,216,217$ temporal relations, $57,992$ causal relations, and $15,841$ subevent relations, which is larger than existing datasets of all the ERE tasks by at least an order of magnitude. Experiments show that ERE on \ourdata is quite challenging, and considering relation interactions with joint learning can improve performances. The dataset and source codes can be obtained from \url{https://github.com/THU-KEG/MAVEN-ERE}.
\end{abstract}

\section{Introduction}

Communicating events is a central function of human languages, and understanding the complex relationships between events is essential to understanding events~\citep{levelt1993speaking,miller2013language,pinker2013learnability}. Thus event relation extraction (ERE) tasks, including extracting event coreference, temporal, causal and subevent relations~\citep{Liu2020ExtractingEA}, are fundamental challenges for natural language processing (NLP) and also support various applications~\citep{chaturvedi-etal-2017-story,rashkin-etal-2018-event2mind,Khashabi2018QA,sap2019atomic,zhang2020TransOMCS}.

\begin{table}[!t]
\small
\centering
\begin{adjustbox}{max width=1\linewidth}
{
\begin{tabular}{lrrrrrr}
\toprule
Dataset   & \#Doc. & \#Event & Coref. & \#T-Link & \#C-Link & \#Subevent   \\ \midrule
ACE 2005  & $599$   & $4,090$     & \mycheck  & \myx   & \myx  & \myx  \\
TAC KBP   & $1,075$   & $19,257$ & \mycheck  & \myx   & \myx  & \myx \\
TB-Dense & $36$   & $1,712$     & \myx  & \multicolumn{1}{r}{$10,750$}  & \myx  & \myx \\
MATRES  & $275$   & $11,861$     & \myx  & \multicolumn{1}{r}{$13,573$}  & \myx  & \myx \\
Causal-TB & $183$   & $6,811$     & \myx  & $5,118$   & \multicolumn{1}{r}{$318$}  & \myx \\
EventStoryLine  & $258$ & $4,732$   & \mycheck  & \multicolumn{1}{r}{$8,111$}   & \multicolumn{1}{r}{$4,584$}  & \myx \\
HiEve  & $100$   & $2,734$     & \mycheck  & \myx   & \myx  & \multicolumn{1}{r}{$3,648$} \\
RED & $95$  & $2,049$   & \mycheck  & \multicolumn{1}{r}{$4,209$}   & \multicolumn{1}{r}{$1,147$}  & \multicolumn{1}{r}{$729$}   \\ 
\midrule
\ourdata  & $4,480$  & $103,193  $   & \mycheck  & $1,216,217$   & $57,992$  & $15,841$   \\ \bottomrule
\end{tabular}
}
\end{adjustbox}
\caption{Comparisons between \ourdata and most widely-used event relation datasets. T-Link and C-Link denote temporal and causal relations, respectively. \#Event is the number of events (coreference chains) for the datasets with coreference annotation; otherwise it is the number of event mentions.}
\label{tab:dataset_motivating}
\end{table}

\begin{figure*}[!t]
    \centering
    \includegraphics[width=\linewidth]{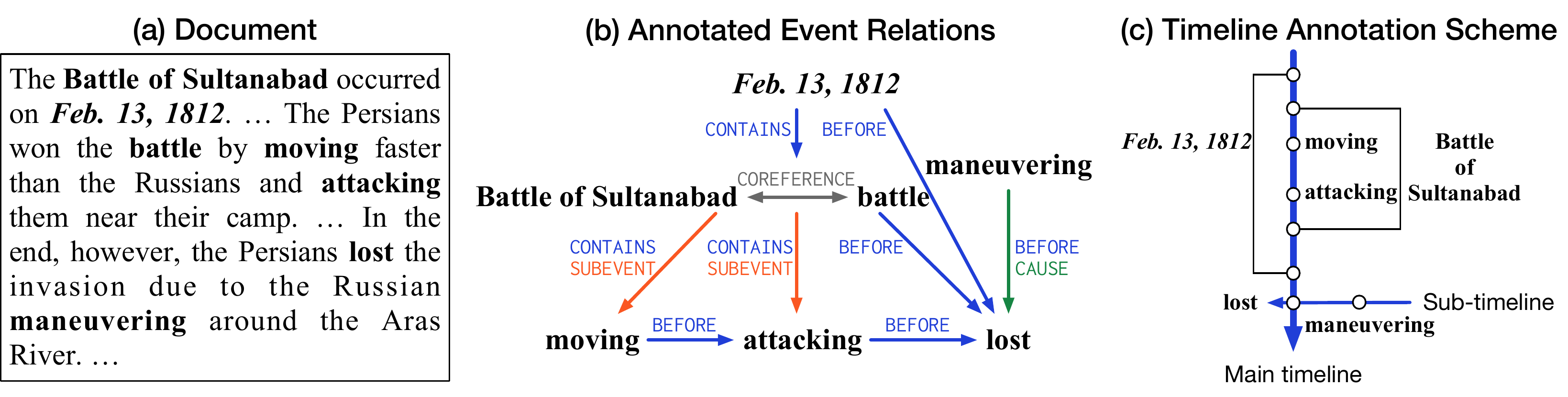}
    \caption{An example document (a) and its annotation results (b) of selected events. \textbf{Bold:} event trigger. \textit{\textbf{Italic bold:}} TIMEX. We also show an instance of annotating its temporal relations with our timeline annotation scheme (c).}
    \label{fig:illustration}
\end{figure*}

Due to the widely acknowledged importance, many efforts have been devoted to developing advanced ERE methods~\citep{liu-etal-2014-supervised,hashimoto-etal-2014-toward,ning-etal-2017-structured}. Recently, data-driven neural models have become the mainstream of ERE methods~\citep{dligach-etal-2017-neural,aldawsari-finlayson-2019-detecting,liu2020knowledge,lu-ng-2021-constrained}. However, these data-driven methods are severely limited by two drawbacks of existing event relation datasets: (1) \textbf{Small data scale}. Due to the high inherent annotation complexity, the data scale of existing human-annotated datasets is limited. From the statistics shown in \cref{tab:dataset_motivating}, we can see existing popular datasets contain only hundreds of documents and limited numbers of relations, which cannot adequately cover the diverse event semantics and is insufficient for training sophisticated neural models~\citep{wang-etal-2020-joint}. Moreover, event relations in these datasets are often incomprehensive. For instance, TB-Dense~\citep{chambers-etal-2014-dense} and MATRES~\citep{ning2018matres} only annotate event temporal relations for event pairs within adjacent sentences. 
(2) \textbf{Absence of unified annotation}. Naturally, various types of event relations have rich interactions with each other. For example, the cause events must start temporally before the effect events, and the superevents must temporally contain the subevents. The coreference relation is the foundation, and all the other relations are shared among coreferent event mentions. However, as shown in \cref{tab:dataset_motivating}, existing datasets typically only cover limited relation types at once. RED~\citep{ogorman2016richer} is a notable exception developing comprehensive unified annotation guidelines, but can only serve as a test set~\citep{wang-etal-2020-joint} due to its small scale. This results in the closely connected ERE tasks being conventionally handled independently and limits the development of joint ERE methods~\citep{ning-etal-2018-joint,wang-etal-2020-joint}.

In this paper, we construct \ourdata, the first unified large-scale event relation dataset, based on the previous MAVEN~\citep{wang-etal-2020-maven} dataset, which is a massive general-domain event detection dataset covering $4,480$ English Wikipedia documents and $168$ fine-grained event types. As the example in \cref{fig:illustration}, \ourdata makes up the absence of unified annotation by annotating $4$ kinds of event relations in the same documents.  \ourdata has $103,193$ event coreference chains, $1,216,217$ temporal relations, $57,992$ causal relations, and $15,841$ subevent relations. To our knowledge, \ourdata achieves the first million-scale human-annotated ERE dataset. As shown in \cref{tab:dataset_motivating}, in every ERE task, \ourdata is larger than existing datasets by at least an order of magnitude, which shall alleviate the limitation of data scale and facilitate developing ERE methods.

As shown in \cref{fig:illustration}, event relations are dense and complex. Hence constructing \ourdata requires thorough and laborious crowd-sourcing annotation. To ensure affordable time and resource costs, we further develop a new annotation methodology based on~\citet{ogorman2016richer}, which is the only existing annotation scheme supporting all the relation types. Specifically, we decompose the overall annotation task into multiple sequential stages, which reduces competence requirements for annotators. The overhead of later stages can also be reduced with the results of previous stages. First, we annotate coreference relations so that the later-stage annotations only need to consider one of all the coreferent event mentions. For temporal relation annotation, we develop a new timeline annotation scheme, which avoids laboriously identifying temporal relations for every event pair like previous works~\citep{chambers-etal-2014-dense,ning2018matres}. This new scheme brings much denser annotation results. For every $100$ words, \ourdata has more than $6$ times the number of temporal relations as the previous most widely-used dataset MATRES~\citep{ning2018matres}. For causal and subevent relation annotation, we set annotation constraints with temporal relations and the relation transitivity to reduce annotation scopes.

We develop strong baselines for \ourdata based on a widely-used sophisticated pre-trained language model~\citep{liu2019roberta}. Experiments show that: (1) ERE tasks are quite challenging and achieved performances are far from promising; (2) Our large-scale data sufficiently trains the models and brings performance benefits; (3) Considering the relation interactions with straightforwardly joint training improves the performances, which encourages more explorations. We also provide some empirical analyses to inspire future works.

\section{Dataset Construction}
Based on the event triggers in MAVEN~\citep{wang-etal-2020-maven}, we annotate data for four ERE tasks: extracting event coreference, temporal, causal, and subevent relations. For each task, we introduce its definition, the annotation process, and basic statistics of \ourdata compared with its typical existing datasets. For the overall statistic comparisons, please refer to \cref{sec:app_size}.

\subsection{Coreference Relation}
\paragraph{Task Description}
Event coreference resolution requires identifying the \textit{event mentions} referring to the same event. Event mentions are the key texts expressing the occurrences of events. For example, in \cref{fig:illustration}, the ``Battle of Sulatnabad'' and the later ``battle'' are two event mentions referring to the same real-world event, so they have a coreference relation. Like entity coreference resolution, event coreference resolution is important to various applications and is widely acknowledged as more challenging~\citep{choubey-huang-2018-improving}. 

\paragraph{Annotation}
We follow the annotation guidelines of~\citet{ogorman2016richer} and invite $29$ annotators to annotate event coreference relations. The annotators are all trained and pass a qualification test before annotation. Given the documents and highlighted event mentions, the annotators are required to group the coreferent mentions together. The outputs are \textit{event coreference chains}, each linking a set of different event mentions. Each document is annotated by $3$ independent annotators, and the final results are obtained by majority voting. To improve the data quality on top of the original MAVEN and avoid annotation vagueness, we allow the annotators to report if the provided mentions do not express events, and we will delete the mentions reported by all the annotators. The B-Cubed F-1~\citep{bagga-baldwin-1998-entity-based} between each pair of annotation results is $91\%$ on average, which shows that the annotation consistency is satisfactory.
\begin{table}[!t]
\small
\centering
\begin{adjustbox}{max width=1\linewidth}
{
\begin{tabular}{lrrrc}
\toprule
Dataset   & \#Doc. & \#Mention & \#Chain & Event Type   \\ \midrule
ACE 2005  & $599$                    & $5,349$                      & $4,090$                    & \mycheck   \\
ECB+      & $982$  & $14,884$  & $9,875$                   & \myx \\

TAC KBP   & $1,075$                  & $29,471$                     & $19,257$                   & \mycheck   \\
\ourdata  & $4,480$                  & $112,276$                    & $103,193$                  & \mycheck   \\ \bottomrule
\end{tabular}
}
\end{adjustbox}
\caption{Statistics about event coreference relations of \ourdata and existing widely-used datasets.}
\label{tab:coref_size}
\end{table}
\paragraph{Statistics} After annotation, we get $103,193$ event coreference chains in total. In \cref{tab:coref_size}, we compare the size of \ourdata with existing widely-used datasets, including ACE 2005~\citep{walker2006ace}, ECB+~\citep{cybulska-vossen-2014-using}, and TAC KBP. Following the setup of previous works~\citep{lu-ng-2021-constrained,lu-ng-2021-conundrums}, the TAC KBP here includes LDC2015E29, LDC2015E68 and TAC KBP 2015~\citep{ellis2015overview}, 2016~\citep{ellis2016overview} and 2017~\citep{getman2017overview}. We can see that \ourdata has much more annotated event coreference chains, which shall benefit event coreference resolution methods.

\subsection{Temporal Relation}
\paragraph{Task Description}

Temporal relation extraction aims at extracting the temporal relations between events and \textit{temporal expressions} (TIMEXs). TIMEXs are the definitive references to time within texts. Considering them in temporal relation extraction helps to anchor the relative temporal orders to concrete timestamps. Hence we need to annotate TIMEXs before annotating temporal relations. 

\looseness=-1 Following the ISO-TimeML standard~\citep{pustejovsky-etal-2010-iso}, we annotate four types of TIMEX: \texttt{DATE}, \texttt{TIME}, \texttt{DURATION} and \texttt{PREPOSTEXP}, but we ignore the \texttt{QUANTIFIER} and \texttt{SET}, since they are harder for crowd-sourcing workers and less helpful for linking events to real-world timestamps. For temporal relations, we follow \citet{ogorman2016richer} and comprehensively set $6$ types of temporal relations: \texttt{BEFORE}, \texttt{CONTAINS}, \texttt{OVERLAP}, \texttt{BEGINS-ON}, \texttt{ENDS-ON}, \texttt{SIMULTANEOUS}. Except for \texttt{SIMULTANEOUS} and \texttt{BEGINS-ON}, the relation types are unidirectional, i.e., the head event must start before the tail event in a relation instance. 

\paragraph{Annotation}
In TIMEX annotation, we invite $112$ trained and qualified annotators. Each document is annotated by $3$ annotators, and the final results are obtained through majority voting. The average inter-annotator agreement is $78.4\%$ (Fleiss' kappa). 

Previous works~\citep{styler-iv-etal-2014-temporal,chambers-etal-2014-dense,ning2018matres} show that annotating temporal relations is very challenging since densely annotating relations for every event pair is extremely time-consuming, and the expressions of temporal relations are often vague. Hence we design a sophisticated annotation scheme inspired by the multi-axis scheme of \citet{ning2018matres} and the time-anchoring scheme of \citet{reimers-etal-2016-temporal}. As illustrated in \cref{fig:illustration} (c), instead of identifying relations for every single event pair, we ask the annotators to sort the beginnings and endings of events and TIMEXs on a \textit{timeline}. Thus the annotators only need to consider how to arrange the bounding points of temporally close events and TIMEXs, and the relations between the events and TIMEXs on the timeline can be automatically inferred from their relative positions. However, due to the narrative vagueness, the temporal relations between some events cannot be clearly determined from contexts, such as the ``maneuvering'' and ``attacking'' in \cref{fig:illustration}. As discussed by~\citet{ning2018matres}, this often happens when expressing opinions, intentions, and hypotheses. In these cases, we allow the annotators to create \textit{sub-timelines}, and we treat events on different timelines as no temporal relations. An event may be placed on multiple timelines like the ``lost'' in \cref{fig:illustration}. 

With this annotation scheme, we can get high-quality temporal relations for all the pairs at an affordable cost with no need to reduce the annotation scope like previous works~\citep{chambers-etal-2014-dense,ning2018matres} which only annotate events within adjacent sentences. To control data quality and resource costs, each document will be annotated by a well-trained annotator at first. Then an expert will check and revise the annotation results. We invite $49$ annotators and $17$ experts in temporal relation annotation. To measure data quality, we randomly sample $100$ documents and annotate them twice in the above pipeline. The average agreement is $67.8\%$ (Cohen's kappa).

\paragraph{Statistics}

We obtain $25,843$ TIMEXs, including $20,654$ \texttt{DATE}, $4,378$ \texttt{DURATION}, $793$ \texttt{TIME}, and $18$ \texttt{PREPOSTEXP}. Based on the events and TIMEXs, we annotate $1,216,217$ temporal relations in total, including $1,042,709$ \texttt{BEFORE}, $152,702$ \texttt{CONTAINS}, $9,937$ \texttt{SIMULTANEOUS}, $9,850$ \texttt{OVERLAP}, $639$ \texttt{BEGINS-ON}, and $380$ \texttt{ENDS-ON}. We can see the data unbalance among types is serious. To ensure that the created dataset well reflects the real-world data distribution, we do not intervene the label distribution and keep the unbalanced distribution in \ourdata. This poses a challenge for future temporal relation extraction models.

\looseness=-1 In \cref{tab:temp_size}, we compare the size of \ourdata with existing widely used datasets, including TimeBank 1.2~\citep{pustejovsky2003timebank}, TempEval 3~\citep{uzzaman-etal-2013-tempeval3}, RED~\citep{ogorman2016richer}, TB-Dense~\citep{chambers-etal-2014-dense}, MATRES~\citep{ning2018matres}, and TCR~\citep{ning-etal-2018-joint}. \ourdata is orders of magnitude larger than existing datasets and is the first million-scale temporal relation extraction dataset to our knowledge. Our timeline annotation scheme also brings denser annotation results. For every $100$ words, \ourdata has $95.3$ temporal relations, while MATRES has $14.3$. We believe a leap in data size could significantly facilitate temporal relation extraction research and promote broad temporal reasoning applications.
\begin{table}[!t]
\small
\centering
\begin{adjustbox}{max width=1\linewidth}
{
\begin{tabular}{lrrrrr}
\toprule
Dataset   & \#Doc. & \#Mention & \#TIMEX  & \#T-Link  \\ \midrule 
TimeBank 1.2  & $183$ & $7,935$  & $1,414$  & $6,115$  \\
TempEval-3$^*$  & $2,472$ & $82,061$  & $15,349$  & $113,848$  \\
RED & $95$ & $8,731$  & $893$  & $4,209$ \\
TB-Dense  & $36$ & $1,712$  & $253$  & $10,750$\\
MATRES   & $275$ & $11,861$  & $1,955$  & $13,573$  \\
TCR & $25$ & $1,134$  & $217$  & $2,660$  \\
\ourdata  & $4,480$ & $112,276$  & $25,843$  & $1,216,217$  \\ \bottomrule
\end{tabular}
}
\end{adjustbox}
\caption{Statistics about temporal relations (T-Links) of \ourdata and existing widely-used datasets. $^*$: the majority of TempEval-3 is automatically annotated.}
\label{tab:temp_size}
\end{table}

\subsection{Causal Relation}

\paragraph{Task Description}
Understanding causality is a long-standing goal of artificial intelligence. Causal relation extraction, which aims at extracting the causal relations between events, is an important task to evaluate it. To enable crowd-sourcing annotation, we do not adopt the complicated causation definitions~\citep{dunietz-etal-2017-because} but instead annotate two types of straightforward and clear causal relation types: \texttt{CAUSE} and \texttt{PRECONDITION} following previous discussions~\citep{ikuta-etal-2014-challenges,ogorman2016richer}. \texttt{CAUSE} is defined as ``the tail event is inevitable given the head event'', and \texttt{PRECONDITION} is defined as ``the tail event would not have happened if the head event had not happened''~\citep{ikuta-etal-2014-challenges}. Note that we allow to annotate causal relations for negative events, which are the events that did not actually happen. In this way, we also cover the negative causation discussed in previous literatures~\citep{mirza-etal-2014-annotating}.

\paragraph{Annotation}
Considering the temporal nature of causality, we limit the annotation scope to event pairs with \texttt{BEFORE} and \texttt{OVERLAP} relations labeled in temporal annotation. To further reduce annotation overhead, we ask the annotators to consider the transitivity of causal relations and make minimal annotations. That is if ``A \texttt{CAUSE/PRECONDITION} B'' and ``B \texttt{CAUSE/PRECONDITION} C'' have been annotated, the causal relation between A and C can be discarded. Furthermore, we annotate causal relations and subevent relations in the same stage so that we can involve subevent relations in the transitivity rules. This means that you can discard the causal relations between A and C if you have (1) ``A \texttt{CAUSES/PRECONDITIONS} B and C \texttt{SUBEVENT} B'' or (2) ``A \texttt{SUBEVENT} B and B \texttt{PRECONDITION} C''. The discarded relations are then automatically completed after human annotation. We invite $58$ trained and qualified annotators, and each document is annotated by $3$ independent annotators. The final results are obtained through majority voting. The average inter-annotator agreement for causal relations is $69.5\%$ (Cohen's kappa).

\begin{table}[!t]
\small
\centering
\begin{adjustbox}{max width=1\linewidth}
{
\begin{tabular}{lrrrrr}
\toprule
Dataset   & \#Doc. & \#Mention & \#C-Link  \\ \midrule
BECauSE 2.0  & $121$ & $1,803$  & $110$  \\
CaTeRS & $320$ & $2,708$  & $488$   \\
RED& $95$ & $8,731$  & $1,147$   \\
Causal-TB  & $183$ & $6,811$  & $318$  \\
EventStoryLine   & $258$ & $4,732$  & $4,584$   \\
\ourdata  & $4,480$ & $112,276$  & $57,992$   \\ \bottomrule
\end{tabular}
}
\end{adjustbox}
\caption{Statistics about causal relations (C-Links) of \ourdata and existing widely-used datasets.}
\label{tab:causal_size}
\end{table}

\looseness=-1 \paragraph{Statistics}
We obtain $57,992$ causal relations, including $10,617$ \texttt{CUASE} and $47,375$ \texttt{PRECONDITION}. \cref{tab:causal_size} shows the size of \ourdata and existing widely-used datasets, including BECauSE 2.0~\citep{dunietz-etal-2017-because}, CaTeRS~\citep{mostafazadeh-etal-2016-caters}, RED~\citep{ogorman2016richer}, Causal-TB~\citep{mirza-etal-2014-annotating}, and EventStoryLine~\citep{caselli-vossen-2017-event}. \ourdata is still much larger than all the existing datasets.

\subsection{Subevent Relation}

\looseness=-1 \paragraph{Task Description}
Subevent relation extraction requires identifying whether event A is a subevent of event B. ``A \texttt{SUBEVENT} B'' means that A is a component part of B and spatiotemporally contained by B~\citep{hovy-etal-2013-events,glavas-etal-2014-hieve,ogorman2016richer}. Subevent relations organize the unconnected events into hierarchical structures, which support the event understanding applications~\citep{aldawsari-finlayson-2019-detecting}.

\paragraph{Annotation}
\looseness=-1 We limit the annotation scope to event pairs with \texttt{CONTAINS} relations considering the inherent temporal containment property in subevent definition. This significantly reduces annotation overhead. The subevent relation annotation is conducted together with causal relations, and we invite the same $58$ annotators. Each document is annotated by $3$ annotators, and the final results are obtained with majority voting. The average inter-annotator agreement is $75.1\%$ (Cohen's kappa).

\begin{table}[!t]
\small
\centering
\begin{adjustbox}{max width=1\linewidth}
{
\begin{tabular}{lrrrrr}
\toprule
Dataset   & \#Doc. & \#Mention & \#Subevent Relation  \\ \midrule
Intelligence Community  & $100$ & $3,919$  & $4,586$  \\
HiEve & $100$ & $3,185$  & $3,648$   \\
RED& $95$ & $8,731$  & $729$   \\
\ourdata  & $4,480$ & $112,276$  & $15,841$   \\ \bottomrule
\end{tabular}
}
\end{adjustbox}
\caption{Statistics about subevent relations of \ourdata and widely-used datasets. }
\label{tab:sub_size}
\end{table}

\looseness=-1 \paragraph{Statistics}
We get $15,841$ subevent relations after annotation. \cref{tab:sub_size} shows the size comparisons of \ourdata and existing datasets, including the Intelligence Community~\citep{hovy-etal-2013-events}, HiEve~\citep{glavas-etal-2014-hieve} and RED~\citep{ogorman2016richer}. We can see that \ourdata is also significantly larger than existing datasets.
\section{Data Analysis}

\subsection{Distance between Related Events}
\label{sec:ana_dis}
\begin{table}[!t]
\small
\centering
\begin{adjustbox}{max width=1\linewidth}
{
\begin{tabular}{llrrrr}
\toprule
    & Dataset                 & $<50$ (\%)& $50-200$ (\%)& $>200$ (\%)& Avg. \\ \midrule
\multicolumn{1}{l|}{\multirow{3}{*}{Coreference}} & ACE 2005                & $36.4$        & $27.9$  & $35.6$            & $192$ \\
\multicolumn{1}{l|}{}                             & KBP                     & $22.8$        & $26.5$  & $50.7$            & $536$ \\
\multicolumn{1}{l|}{}                             & \ourdata & $31.9$        & $49.4$  & $18.8$            & $122$  \\ \midrule
\multicolumn{1}{l|}{\multirow{4}{*}{Temporal}}    & TB-Dense                & $94.7$        & $5.4$   & $0.0$             & $22$   \\
\multicolumn{1}{l|}{}                             & MATRES                  & $90.8$        & $9.1$   & $0.0$             & $26$   \\
\multicolumn{1}{l|}{}                             & TCR                     & $93.4$        & $6.6$   & $0.0$             & $24$   \\
\multicolumn{1}{l|}{}                             & \ourdata & $27.6$        & $46.9$  & $25.5$            & $147$  \\ \midrule
\multicolumn{1}{l|}{\multirow{3}{*}{Causal}}      & Causal-TB               & $100.0$       & $0.0$  & $0.0$             & $11$   \\
\multicolumn{1}{l|}{}                             & EventStoryLine          & $59.3$        & $32.2$  & $8.4$             & $76$   \\
\multicolumn{1}{l|}{}                             & \ourdata & $48.6$        & $37.6$  & $13.7$            & $92$   \\ \midrule
\multicolumn{1}{l|}{\multirow{2}{*}{Subevent}}    & HiEve                   & $33.1$       & $39.6$  & $27.3$            & $152$  \\
\multicolumn{1}{l|}{}                             & \ourdata & $30.6$       & $50.0$  & $19.4$            & $124$ \\
\bottomrule
\end{tabular}
}
\end{adjustbox}
\caption{The distributions and average values of distances (measured in \#words) between related events of different relation types.}
\label{tab:distance}
\end{table}

Understanding the relations between long-distance event pairs helps to understand documents in the discourse-level~\citep{naik-etal-2019-tddiscourse}, and modeling long-range dependencies is a long-standing challenge for NLP models. Hence we analyze the distance distributions of the annotated event relations in \ourdata and compare them with existing most widely-used datasets in \cref{tab:distance}.

For temporal relations, since the mainstream annotation scheme requires identifying relations for every event pair, existing most widely-used and high-quality datasets like TB-Dense and MATRES limit the annotation scope to the events in the same or adjacent sentences and ignore long-distance temporal relations, which are also informative~\citep{reimers-etal-2016-temporal,naik-etal-2019-tddiscourse}. This also limits the causal relation datasets based on them like Causal-TB. As shown in \cref{tab:distance}, with the help of our timeline annotation scheme, \ourdata has much more long-distance temporal and causal relations compared to existing datasets, which can better support real-world applications and poses new challenges for ERE models.

For coreference relations, \ourdata has shorter average distances and much higher short-distance rates. This is because MAVEN~\citep{wang-etal-2020-maven} covers much more generic events and annotates much denser event mentions. For comparison, \ourdata has $8.8$ event mentions per $100$ words, while this number is $1.8$ and $4.2$ for ACE 2005 and TAC KBP, respectively. For subevent relations, the distributions of HiEve and \ourdata are similar, and we think HiEve has a longer average distance because of its longer average document length ($333$ vs. $284$ words).

\begin{figure}[!t]
    \centering
    \scalebox{1.0}{
    \includegraphics[width=\linewidth]{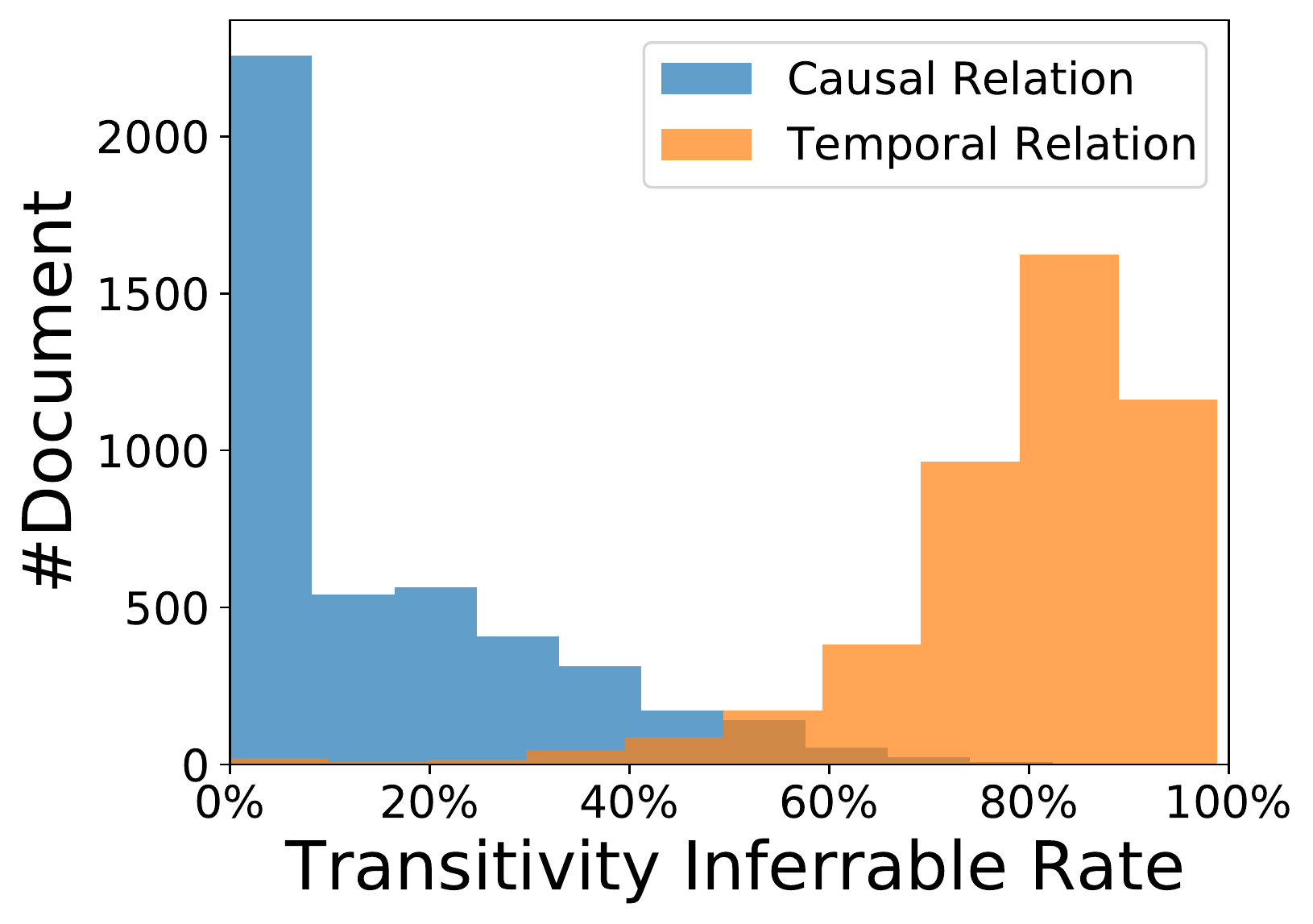}
    }
    \caption{Distribution of documents with different rates of transitivity inferrable temporal and causal relations.}
    \label{fig:trans}
\end{figure}

\subsection{Relation Transitivity}
\label{sec:trans}
Temporal and causal relations follow a certain transitivity rules~\citep{allen1983maintaining,GEREVINI1995207}, e.g., if there exists ``A \texttt{BEFORE} B'' and ``B \texttt{BEFORE} C'', ``A \texttt{BEFORE} C'' also holds. Previous ERE methods often use these natural transitivity rules as constraints in post-processing~\citep{chambers-jurafsky-2008-jointly,denis2011predict,ning-etal-2018-joint} and training~\citep{wang-etal-2020-joint}. Here we estimate the importance of considering transitivity in handling \ourdata by counting how many relations can be inferred from other relations with transitivity rules. The detailed transitivity rules that we consider are shown in \cref{sec:app_transitivity}.

Overall, $88.8\%$ temporal relations and $23.9\%$ causal relations are inferrable with transitivity rules. We further plot the distribution of documents containing different rates of transitivity inferrable relations in \cref{fig:trans}. We can see that more than $60\%$ temporal relations can be inferred with transitivity rules for most of the documents. The transitivity inferrable causal relations, although significantly less, also take up a substantial proportion. These results suggest that considering the relation transitivity is helpful for handling \ourdata, and we encourage future works to explore it.

\section{Experiments and Analyses}
\label{sec:exp}
To demonstrate the challenges of \ourdata and analyze the potential future directions for ERE, we conduct a series of experiments.
\subsection{Experiment Setup}
\paragraph{Model} Considering that pre-trained language models (PLMs) have dominated broad NLP tasks, we adopt a widely-used PLM \RBT~\citep{liu2019roberta} as the backbone and build classification models on top of it, which provides simple but strong baselines for the $4$ ERE tasks. To extract the event relations in a document, we encode the whole document with \RBT and set an additional classification head taking the contextualized representations at the positions of different event pairs' corresponding event triggers as inputs. Then we fine-tune the model to classify relation labels. Besides training the $4$ tasks independently, we also set a straightforward jointly training model combining the losses of the $4$ tasks, which is to demonstrate the benefits of our unified annotation. The implementation details are shown in \cref{sec:app_implement}.

\begin{table*}[!t]
\small
\centering
\begin{adjustbox}{max width=1\linewidth}
{
\begin{tabular}{l|rrr|rrr|rrr|rrr}
\toprule
\multirow{2}{*}{} & \multicolumn{3}{c|}{MUC}                                                              & \multicolumn{3}{c|}{B$^3$}                                                            & \multicolumn{3}{c|}{CEAF$_e$}                                                         & \multicolumn{3}{c}{BLANC}                                                            \\ \cmidrule{2-13} 
                  & \multicolumn{1}{c}{Precision} & \multicolumn{1}{c}{Recall} & \multicolumn{1}{c|}{F-1} & \multicolumn{1}{c}{Precision} & \multicolumn{1}{c}{Recall} & \multicolumn{1}{c|}{F-1} & \multicolumn{1}{c}{Precision} & \multicolumn{1}{c}{Recall} & \multicolumn{1}{c|}{F-1} & \multicolumn{1}{c}{Precision} & \multicolumn{1}{c}{Recall} & \multicolumn{1}{c}{F-1} \\ \midrule
ACE 2005  & $79.1_{1.66}$ & $74.2_{2.88}$ & $76.5_{1.90}$ & $93.1_{0.52}$ & $90.7_{1.16}$ & $91.9_{0.57}$ & $87.2_{1.06}$ & $89.4_{0.39}$ & $88.3_{0.50}$ & $84.8_{1.54}$ & $81.3_{2.80}$ & $82.9_{2.01}$  \\
TAC KBP   & $69.6_{1.58}$ & $74.6_{2.03}$ & $72.0_{0.33}$ & $84.7_{1.49}$ & $88.9_{0.77}$ & $86.7_{0.43}$ & $85.4_{0.61}$ & $82.4_{1.57}$ & $83.9_{0.55}$ & $75.8_{1.15}$ & $82.1_{0.84}$ & $78.5_{0.54}$  \\
\ourdata  & $79.2_{2.20}$ & $\textbf{84.0}_{1.78}$ & $81.4_{0.51}$ & $97.7_{0.35}$ & $\textbf{98.4}_{0.17}$ & $98.1_{0.10}$ & $\textbf{98.0}_{0.11}$ & $97.5_{0.35}$ & $97.7_{0.13}$ & $87.9_{1.18}$ & $\textbf{92.0}_{0.83}$ & $89.8_{0.36}$  \\
\quad+joint  & $\textbf{81.4}_{1.64}$ & $82.8_{1.56}$ & $\textbf{82.1}_{0.43}$ & $\textbf{98.0}_{0.27}$ & $98.3_{0.18}$ & $\textbf{98.2}_{0.11}$ & $\textbf{98.0}_{0.13}$ & $\textbf{97.8}_{0.21}$ & $\textbf{97.9}_{0.09}$ & $\textbf{88.8}_{1.05}$ & $91.4_{1.15}$ & $\textbf{90.2}_{0.27}$  \\\bottomrule
\end{tabular}
}
\end{adjustbox}
\caption{Event coreference resolution performances (\%) of \RBT on \ourdata and existing datasets. We report averages and standard deviations over $5$ random trials. ``+joint'' denotes jointly training on $4$ ERE tasks. \textbf{Bold} denotes higher values among the two results on \ourdata.}
\label{tab:exp_coref}
\end{table*}

\begin{table}[!t]
\small
\centering
\begin{adjustbox}{max width=1\linewidth}
{
\begin{tabular}{llrrrrr}
\toprule
    &               & \multicolumn{1}{c}{Precision} & \multicolumn{1}{c}{Recall} & \multicolumn{1}{c}{F-1} \\ \midrule
\multicolumn{1}{l|}{\multirow{5}{*}{Temporal}}    & TB-Dense & $64.2_{2.12}$ & $49.3_{2.12}$ & $55.8_{1.51}$  \\\multicolumn{1}{l|}{}                             & MATRES  & $75.5_{1.50}$ & $83.8_{1.21}$ & $79.4_{0.64}$  \\\multicolumn{1}{l|}{}                             & TCR   & $84.8_{0.96}$ & $81.1_{2.11}$ & $82.9_{0.74}$  \\\multicolumn{1}{l|}{}                             & \ourdata & $\textbf{57.8}_{0.73}$ & $53.9_{1.36}$ & $55.8_{0.42}$  \\
\multicolumn{1}{l|}{}                             & \quad+joint & $55.4_{0.91}$  & $\textbf{56.6}_{1.52}$& $\textbf{56.0}_{0.59}$ \\
 \midrule
\multicolumn{1}{l|}{\multirow{4}{*}{Causal}}      & Causal-TB  & $50.4_{6.65}$ & $5.9_{0.53}$ & $10.0_{0.82}$  \\\multicolumn{1}{l|}{}                         & EventStoryLine  & $31.1_{1.94}$ & $10.7_{0.88}$ & $14.4_{0.94}$ \\
\multicolumn{1}{l|}{}                             & \ourdata & $\textbf{35.0}_{0.72}$ & $27.2_{0.76}$ & $30.6_{0.44}$  \\
\multicolumn{1}{l|}{}                             & \quad+joint & $33.8_{1.00}$ & $\textbf{29.5}_{0.83}$ & $\textbf{31.5}_{0.42}$ \\
\midrule
\multicolumn{1}{l|}{\multirow{3}{*}{Subevent}}    & HiEve  & $20.0_{1.21}$ & $16.0_{1.16}$ & $17.8_{1.13}$  \\
\multicolumn{1}{l|}{}                             & \ourdata & $29.6_{1.99}$ & $24.6_{3.02}$ & $26.7_{1.34}$  \\
\multicolumn{1}{l|}{}                             & \quad+joint & $\textbf{29.8}_{1.76}$ & $\textbf{25.6}_{1.57}$  & $\textbf{27.5}_{1.10}$ \\

\bottomrule
\end{tabular}
}
\end{adjustbox}
\caption{Performances (\%) of \RBT for extracting temporal, causal, and subevent relations on \ourdata and existing datasets.} 
\label{tab:exp_rel}
\end{table}

\looseness=-1 \paragraph{Benchmarks} To assess the challenges of \ourdata, we also include existing most widely-used datasets of the $4$ ERE tasks into evaluations, including ACE 2005, TAC KBP, TB-Dense, MATRES, TCR, Causal-TB, EventStoryLine, and HiEve. Following previous works~\citep{ning-etal-2018-joint}, TCR is used only as an additional test set for models developed on MATRES. Due to the small data scale of Causal-TB and EventStoryLine, previous works~\citep{gao-etal-2019-modeling,cao-etal-2021-knowledge} typically adopt $5$-fold cross-validation on them and only do causality identification, which ignores the directions of causal relations. In our evaluation on the two datasets, we also do cross-validation but consider the relation directions in accordance with \ourdata. Similarly, we do not down-sample the negative instances for HiEve like previous works~\citep{zhou-etal-2020-temporal,WZCR21}. For the other datasets, we follow previous benchmark settings and show detailed data split statistics in \cref{sec:app_split}.

\paragraph{Metrics} Following previous works~\citep{choubey-huang-2017-event,lu2022e2e}, we adopt MUC~\citep{vilain-etal-1995-model}, B$^3$~\citep{bagga-baldwin-1998-entity-based}, CEAF$_e$~\citep{luo-2005-coreference} and BLANC~\citep{recasense2011blanc} metrics for event coreference resolution. For the other $3$ tasks, we adopt the standard micro-averaged precision, recall and F-1 metrics.

\subsection{Experimental Result}

Experimental results for coreference relations are shown in \cref{tab:exp_coref} and for the other $3$ ERE tasks are shown in \cref{tab:exp_rel}. We can observe that: (1) For extracting coreference, causal and subevent relations, the model's performances on \ourdata are much higher than on previous datasets, indicating the benefits of our larger data scale. (2) For temporal relations, the performances on MATRES and TCR are significantly higher than that on \ourdata. This is because they only cover $4$ relation types and annotate local event pairs within adjacent sentences, which results in easier data and inflated model performances. With the timeline annotation scheme, \ourdata annotates $6$-type global temporal relations within documents, and the lower performance better reflects the inherent challenge of temporal understanding. The performance on TB-Dense is much lower, but we think this comes from TB-Dense's small data scale ($36$ documents), which cannot well train the model. (3) Except for coreference, the achieved performances for the other $3$ ERE tasks are far from practically usable. This demonstrates that understanding the diverse and complex event relations is a huge challenge for NLP models and needs more research efforts. (4) Straightforwardly joint training on the $4$ tasks can bring certain improvements, especially on the tasks with fewer data, i.e., causal and subevent ERE. It indicates that considering the rich interactions between event relations is promising for handling the complex ERE tasks.

\subsection{Analysis on Data Scale}
\begin{figure}[!t]
    \centering
    \scalebox{1.0}{
    \includegraphics[width=\linewidth]{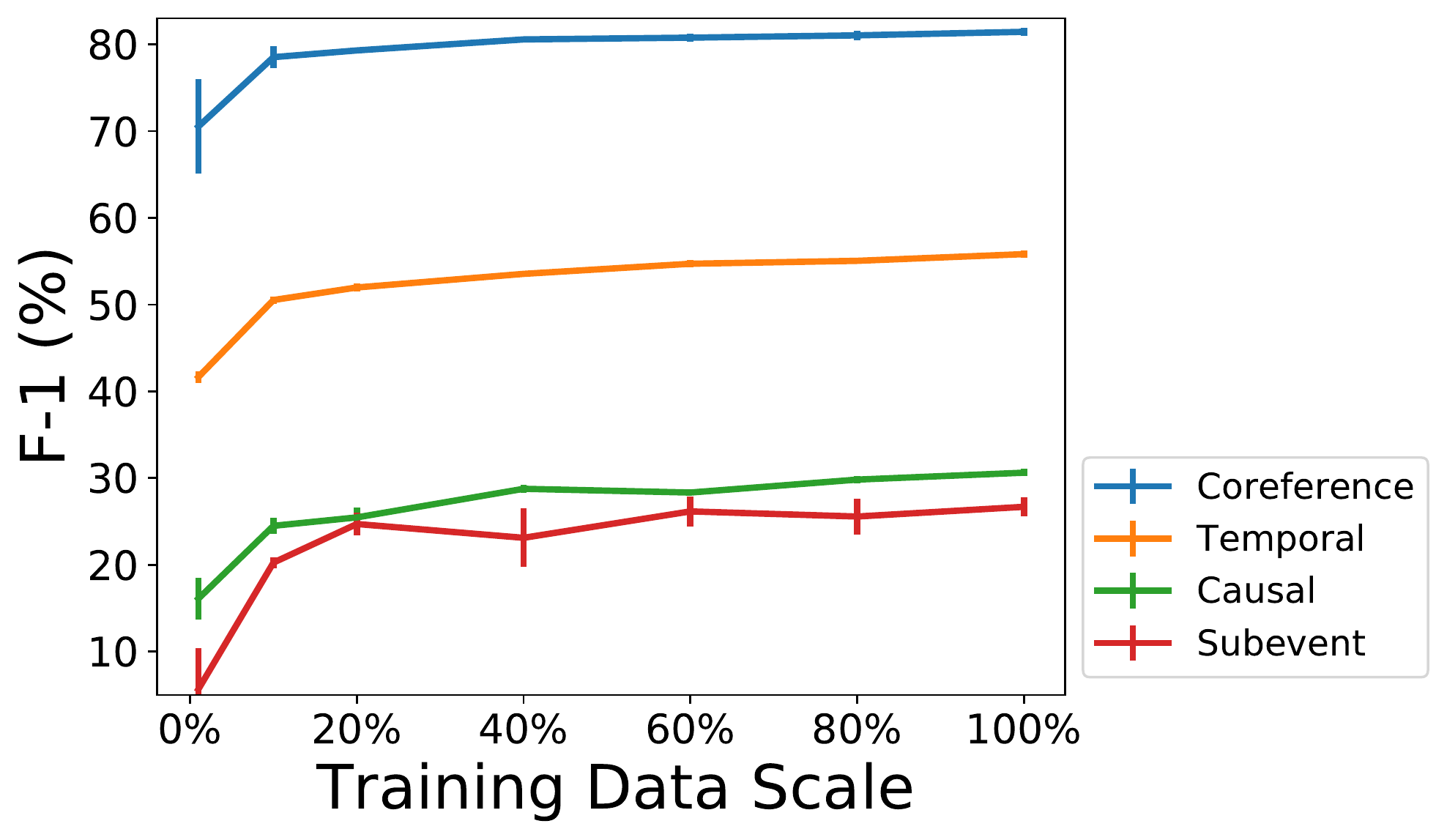}
    }
    \caption{\RBT test results (F-1, MUC for coreference) change along with the training data scale. Error bars indicate standard deviations over $5$ runs.}
    \label{fig:data_ablation}
\end{figure}

Compared with existing datasets, \ourdata significantly increases the data scale of all the ERE tasks. To assess the benefits brought by larger data scale and evaluate whether \ourdata provides enough training data, we conduct an ablation study on the training data scale. 

\cref{fig:data_ablation} shows how \RBT's test performance changes along with different proportions of data used in training. We can see that increasing training data scale brings substantially higher and stabler performances, which shows the benefits of \ourdata's large scale. The performance improvements are quite marginal at the scale of \ourdata. It indicates that \ourdata is generally sufficient to train ERE models.

\subsection{Analysis on Distance between Events}

Like \cref{sec:ana_dis}, we analyze how the distances between related events influence model performances. We sample a jointly-trained model and see how it performs on data with different distances in \cref{tab:exp_distance}. Since the evaluation of event coreference resolution is based on clusters, which cannot be divided by distances, we only study the other $3$ tasks here.

For causal and subevent relations, performances on data with longer distances are lower, which intuitively suggests that modeling long-range dependency is still important to ERE, although the PLMs are effective. However, for temporal relations, data with longer distances are easier. We think this is because event pairs with longer narrative distances are typically also with longer temporal distances, which makes their relations easier to classify.

\subsection{Error Analysis}

\begin{table}[!t]
\small
\centering
\begin{adjustbox}{max width=1\linewidth}
{
\begin{tabular}{lrrr}
\toprule
   & $<50$ & $50-200$ & $>200$  \\ \midrule
Temporal & $52.9$  & $56.8$  & $58.6$   \\
Causal  & $32.7$ & $31.5$  & $29.1$     \\
Subevent & $29.0$ & $28.3$ & $26.6$     \\ 
\bottomrule
\end{tabular}
}
\end{adjustbox}
\caption{\RBT performance (F-1, \%) on data groups with different distances (measured in \#words) between related events of different relation types.}
\label{tab:exp_distance}
\end{table}

We further analyze the errors in the predictions of a jointly trained model to provide insights for further improvements. Considering the event coreference resolution task has reached a high performance and its different cluster-based evaluation, we only analyze the other $3$ tasks. The results are shown in \cref{tab:exp_error}. We can see that identification mistakes (false positive and false negative) make up the majority of all the mistakes. It indicates that the most important challenge for ERE is still identifying whether there is a relation or not. Furthermore, like \cref{sec:trans}, we analyze how many mistakes can be fixed by applying transitivity rules to other predictions. These transitivity fixable mistakes only account for small proportions, which suggests that sophisticated models have imperfectly but substantially learned the transitivity rules from massive data.

\section{Related Work}

Since the fundamental role of understanding event relations in NLP, various ERE datasets have been constructed. Event coreference relations are often covered in event extraction datasets like MUC~\citep{grishman-sundheim-1996-message}, ACE~\citep{walker2006ace} and TAC KBP~\citep{ellis2015overview,ellis2016overview,getman2017overview}. Besides, some datasets focus on unrestricted coreference resolution and ignore event semantic types, like OntoNotes~\citep{Pradhan2007UnrestrictedCI} and ECB datasets~\citep{bejan-harabagiu-2008-linguistic,lee-etal-2012-joint,cybulska-vossen-2014-using}. Following the TimeML specification~\citep{pustejovsky2003timeml,pustejovsky-etal-2010-iso}, established temporal relation datasets like TimeBank~\citep{pustejovsky2003timebank} and TempEval~\citep{verhagen2009tempeval,verhagen-etal-2010-tempeval2,uzzaman-etal-2013-tempeval3} have been constructed. However, these works exhibit low annotation agreements and efficiency issues. \citet{ning2018matres} develop a multi-axis annotation annotation scheme based on the dense scheme of \citet{chambers-etal-2014-dense} to alleviate them, and \citet{reimers-etal-2016-temporal} propose to anchor the event starting and ending points to specific time. Our timeline annotation scheme is inspired by them. Based on the temporal understanding, causal relation datasets~\citep{do-etal-2011-minimally,mirza-etal-2014-annotating,mostafazadeh-etal-2016-caters,dunietz-etal-2017-because,caselli-vossen-2017-event,tan-etal-2022-causal} are developed. To organize events into hierarchies, subevent relation datasets~\citep{hovy-etal-2013-events,glavas-etal-2014-hieve} are collected.

\begin{table}[!t]
\small
\centering
\begin{adjustbox}{max width=1\linewidth}
{
\begin{tabular}{lrrr}
\toprule
   & FP & FN & Transitivity Fixable  \\ \midrule
Temporal & $38.78$  & $53.75$  & $0.85$   \\
Causal  & $37.73$ & $59.88$  & $0.23$     \\
Subevent & $48.64$ & $51.36$ & $-$     \\ 
\bottomrule
\end{tabular}
}
\end{adjustbox}
\caption{Rates (\%) of different kinds of mistakes in \RBT predictions. FP denotes false positive. FN denotes false negative.}
\label{tab:exp_error}
\end{table}

However, the scale of these datasets is limited, and different types of relations are rarely integrated into one dataset. Some datasets~\citep{hovy-etal-2013-events,mirza-etal-2014-annotating,glavas-etal-2014-hieve,caselli-vossen-2017-event,minard-etal-2016-meantime,ning-etal-2018-joint} annotate two or three kinds of relations. \citet{ogorman2016richer} and \citet{hong-etal-2016-building} provide unified annotation schemes for within-document and cross-document event relations, respectively, but their constructed datasets are also small. We construct \ourdata referring to the guidelines of \citet{ogorman2016richer}.

\section{Conclusion and Future Work}
We present \ourdata, a unified large-scale dataset for event coreference, temporal, causal, and subevent relations, which significantly alleviates the small scale and absence of unified annotation issues of previous datasets. Experiments show that real-world event relation extraction is quite challenging and may be improved by jointly considering multiple relation types and better modeling long-range dependency. In the future, we will extend the dataset to more scenarios like covering more event-related information and languages.

 \section*{Limitations}

The most important limitation of \ourdata is that it only covers English documents, which is inherited from the original MAVEN~\citep{wang-etal-2020-maven} dataset. This limits the linguistic features covered by \ourdata and the scope of applications built on it. We encourage future works to explore (1) develop models for the low-resource languages by applying multilingual transfer learning techniques to \ourdata; (2) annotate native datasets for the low-resource languages with the annotation schemes of \ourdata. Another limitation is that \ourdata only covers the within-document event relations. Future works may extend \ourdata to cross-document event relations with the help of existing explorations~\citep{cybulska-vossen-2014-using,hong-etal-2016-building}.

\section*{Ethical Considerations}
\looseness=-1 This paper presents a new dataset, and we discuss some related ethical considerations here. (1) \textbf{Intellectual property}. The original MAVEN dataset is shared under the CC BY-SA 4.0 license\footnote{\url{https://creativecommons.org/licenses/by-sa/4.0}} and the Wikipedia corpus is shared under the CC BY-SA 3.0 license\footnote{\url{https://creativecommons.org/licenses/by-sa/3.0}}. They are both free for research use, and we develop \ourdata with the consent of the authors of MAVEN. (2) \textbf{Worker Treatments.} We hire the annotators from multiple professional data annotation companies and fairly pay them with agreed salaries and workloads. All employment is under contract and in compliance with local regulations. (3) \textbf{Controlling Potential Risks.} Since the texts in \ourdata do not involve private information and annotating event relations does not require many judgments about social issues, we believe \ourdata does not create additional risks. To ensure it, we manually checked some randomly sampled data and did not note risky issues.

\section*{Acknowledgements}
This work is supported by the New Generation Artificial Intelligence of China (2020AAA0106501), the Institute for Guo Qiang, Tsinghua University (2019GQB0003), the Pattern Recognition Center, WeChat AI, Tencent Inc, and the Tsinghua University - Siemens Ltd., China Joint Research Center for Industrial Intelligence and Internet of Things. We thank all the annotators for their efforts and the anonymous reviewers for their valuable comments.
\bibliography{anthology,custom}
\bibliographystyle{acl_natbib}

\clearpage
\appendix
\section*{Appendix}
\section{Dataset Statistics Comparison}
\label{sec:app_size}
We show the detailed statistics of \ourdata and other existing widely-used datasets in \cref{tab:all_stat}. For the TAC KBP data, we follow the setup of \citet{lu-ng-2021-conundrums}, which includes TAC KBP 2015~\citep{ellis2015overview}, 2016~\citep{ellis2016overview}, 2017~\citep{getman2017overview} as well as LDC2015E29 and LDC2015E68. For the multilingual datasets, we only report their statistics of the English parts in accordance with \ourdata.

\section{Transitivity Rules}
\label{sec:app_transitivity}
\cref{tab:trans_rules} shows the transitivity rules of temporal and causal relations that we consider in \cref{sec:trans}. ``\texttt{Relation\_1} + \texttt{Relation\_2} = \texttt{Relation\_3}'' means if there exists ``A \texttt{Relation\_1} B'' and ``B \texttt{Relation\_2} C'', ``A \texttt{Relation\_3} C'' also holds.

\begin{table}[h!]
    \centering
    \begin{adjustbox}{max width=1\linewidth}
{
    \begin{tabular}{c}
    \toprule
        Temporal Transitivity Rules \\
        \midrule
\texttt{BEFORE} + \texttt{BEFORE} = \texttt{BEFORE} \\
\texttt{BEFORE} + \texttt{CONTAINS} = \texttt{BEFORE} \\
\texttt{BEFORE} + \texttt{SIMULTANEOUS} = \texttt{BEFORE} \\
\texttt{BEFORE} + \texttt{OVERLAP} = \texttt{BEFORE} \\
\texttt{BEFORE} + \texttt{BEGINS-ON} = \texttt{BEFORE} \\
\texttt{BEFORE} + \texttt{ENDS-ON} = \texttt{BEFORE} \\
\texttt{CONTAINS} + \texttt{CONTAINS} = \texttt{CONTAINS} \\
\texttt{CONTAINS} + \texttt{SIMULTANEOUS} = \texttt{CONTAINS} \\
\texttt{SIMULTANEOUS} + \texttt{SIMULTANEOUS} = \texttt{SIMULTANEOUS} \\
\texttt{SIMULTANEOUS} + \texttt{BEFORE} = \texttt{BEFORE} \\
\texttt{SIMULTANEOUS} + \texttt{CONTAINS} = \texttt{CONTAINS} \\
\texttt{SIMULTANEOUS} + \texttt{OVERLAP} = \texttt{OVERLAP} \\
\texttt{SIMULTANEOUS} + \texttt{BEGINS-ON} = \texttt{BEGINS-ON} \\
\texttt{SIMULTANEOUS} + \texttt{ENDS-ON} = \texttt{ENDS-ON} \\
\texttt{OVERLAP} + \texttt{BEFORE} = \texttt{BEFORE} \\
\texttt{OVERLAP} + \texttt{SIMULTANEOUS} = \texttt{OVERLAP} \\
\texttt{BEGINS-ON} + \texttt{BEGINS-ON} = \texttt{BEGINS-ON} \\
\texttt{BEGINS-ON} + \texttt{SIMULTANEOUS} = \texttt{BEGINS-ON} \\
\texttt{ENDS-ON} + \texttt{CONTAINS} = \texttt{BEFORE} \\
\texttt{ENDS-ON} + \texttt{SIMULTANEOUS} = \texttt{ENDS-ON} \\
\texttt{ENDS-ON} + \texttt{BEGINS-ON} = \texttt{ENDS-ON} \\
\midrule 
Causal Transitivity Rules \\
\midrule 
\texttt{CAUSE} + \texttt{CAUSE} = \texttt{CAUSE} \\
\texttt{CAUSE} + \texttt{PRECONDITION} = \texttt{PRECONDITION} \\
\texttt{PRECONDITION} + \texttt{PRECONDITION} = \texttt{PRECONDITION} \\
    \bottomrule
    \end{tabular}
}
\end{adjustbox}
\caption{Relation transitivity rules considered.}
\label{tab:trans_rules}
\end{table}

\section{Implementation Details}
\label{sec:app_implement}
\looseness=-1 We implement the \RBT model using the Huggingface's Transformers library~\citep{wolf-etal-2020-transformers}. \RBT contains $110$M parameters, and we add a two-layer perceptron with $150$ hidden dimensions and $0.2$ dropout rate as the classification head. We use the standard cross-entropy loss for event temporal, causal, and subevent relation extraction tasks. For event coreference resolution, we follow the design of \citet{joshi-etal-2019-bert}. We use the Adam~\citep{adam} optimizer to train the models and set $200$ warmup steps. For independently trained models, we set the learning rates as $1\times 10^{-4}$ and $1\times 10^{-5}$ for the classification head and the \RBT encoder. For jointly trained models, the learning rates are $3\times 10^{-4}$ and $2\times 10^{-5}$ for the classification head and the encoder, respectively. We set the factors as $0.4$, $2.0$, $4.0$, and $4.0$ for the losses of coreference, temporal, causal, and subevent relations. These hyper-parameters are manually tuned with $10$ runs and selected with F-1 scores. We use a GeForce RTX 3090 GPU to run the experiments. Average runtimes for an experiment are about $2.2$, $2.3$, $1.1$, $0.5$, and $3.4$ hours for coreference ERE, temporal ERE, causal ERE, subevent ERE, and joint training.

In evaluation, we implement the standard precision, recall, and F-1 scores with the scikit-learn toolkit\footnote{\url{https://scikit-learn.org}}. For event coreference resolution, we implement the evaluation metrics referring to \url{https://github.com/kentonl/e2e-coref}.

\section{Data Split Statistics}
\label{sec:app_split}
In all the ERE experiments, we split \ourdata as the original split in \citet{wang-etal-2020-maven}.

In event coreference resolution, for ACE 2005 and TAC KBP data, we follow the split of \citet{lu-ng-2021-conundrums}. For TAC KBP, LDC2015E29, LDC2015E68, TAC KBP 2015, and TAC KBP 2016 are used for training, and TAC KBP 2017 is used for test. The development set is $82$ documents randomly sampled from the training set. However, the data of some LDC catalog numbers provided by \citet{lu-ng-2021-conundrums} are not available, and we use other LDC datasets instead. Specifically, LDC2015E73 and LDC2015E94 are the datasets provided during the TAC KBP 2015 contest and are not publicly available. We use the 2015 data in LDC2020T13 instead. LDC2016E64 and LDC2017E51 are plain source corpora without annotation. We use the 2016 and 2017 data in LDC2020T18 instead. The statistics are shown in \cref{tab:split_coref}.

In event temporal relation extraction, we follow the splits of \citet{ning-etal-2019-improved} and \citet{tan-etal-2021-extracting}. The detailed statistics are shown in \cref{tab:split_temporal}.

In event causal relation extraction, we follow previous works~\citep{gao-etal-2019-modeling,cao-etal-2021-knowledge} to do $5$-fold cross-validation on Causal-TB and EventStoryLine. The statistics for \ourdata are shown in \cref{tab:split_causal}.

In subevent relation extraction, we split HiEve following previous works~\citep{zhou-etal-2020-temporal,WZCR21}. The statistics are shown in \cref{tab:split_subevent}.

\begin{table*}[!t]
\small
\centering
\begin{adjustbox}{max width=1.4\linewidth, angle=270}
{
\begin{tabular}{lrrrrrrrrrrrrr}
\toprule
Dataset & \multicolumn{1}{c}{\#Doc.} & \multicolumn{1}{c}{\#Sentence} & \multicolumn{1}{c}{\#Word} & \multicolumn{1}{c}{\#Event Type} & \multicolumn{1}{c}{\#Mention} & \multicolumn{1}{c}{\#Chain} & \multicolumn{1}{c}{\#TIMEX Type} & \multicolumn{1}{c}{\#TIMEX} & \multicolumn{1}{c}{\#T-Link Type} & \multicolumn{1}{c}{\#T-Link} & \multicolumn{1}{c}{\#C-Link Type} & \multicolumn{1}{c}{\#C-Link} & \multicolumn{1}{c}{\#Subevent Rel.} \\ \midrule
ACE 2005~\citep{walker2006ace} & $599$ & $15,670$ & $294,857$ & $33$ & $5,349$ & $4,090$ & $-$ & $4,960$ & $-$ & $-$ & $-$ & $-$ & $-$ \\
TAC KBP & $1,075$ & $33,208$ & $694,540$ & $38$ & $29,471$ & $19,257$ & $-$ & $-$ & $-$ & $-$ & $-$ & $-$ & $-$ \\
OntoNotes$^\dag$~\citep{Pradhan2007UnrestrictedCI} & $2,384$ & $84,789$ & $1,673,793$ & $-$ & $210,994$ & $47,834$ & $-$ & $-$ & $-$ & $-$ & $-$ & $-$ & $-$ \\
ECB+~\citep{cybulska-vossen-2014-using} & $982$ & $15,812$ & $362,546$ & $14$ & $14,884$ & $9,875$ & $-$ & $-$ & $-$ & $-$ & $-$ & $-$ & $-$ \\

TimeBank 1.2~\citep{pustejovsky2003timebank} & $183$ & $2,611$ & $63,987$ & $7$ & $7,935$ & $-$ & $4$ & $1,414$ & $13$ & $6,115$ & $-$ & $-$ & $-$ \\
TempEval-1~\citep{verhagen2009tempeval} & $183$ & $2,611$ & $63,987$ & $7$ & $7,935$ & $-$ & $4$ & $1,414$ & $6$ & $5,790$ & $-$ & $-$ & $-$ \\
TempEval-2~\citep{verhagen-etal-2010-tempeval2} & $173$ & $2,383$ & $58,214$ & $7$ & $6,158$ & $-$ & $4$ & $1,127$ & $6$ & $4,867$ & $-$ & $-$ & $-$ \\
TempEval-3$^*$~\citep{uzzaman-etal-2013-tempeval3} & $2,472$ & $25,824$ & $672,684$ & $7$ & $82,061$ & $-$ & $4$ & $15,349$ & $13$ & $113,848$ & $-$ & $-$ & $-$ \\
TCR~\citep{ning-etal-2018-joint} & $25$ & $694$ & $17,304$ & $-$ & $1,134$ & $-$ & $3$ & $217$ & $3$ & $2,660$ & $-$ & $-$ & $-$ \\
TB-Dense~\citep{chambers-2013-event} & $36$ & $598$ & $12,543$ & $7$ & $1,712$ & $-$ & $4$ & $253$ & $6$ & $10,750$ & $-$ & $-$ & $-$ \\
MATRES~\citep{ning2018matres} & $275$ &  $2,172$ & $108,999$ & $7$ & $11,861$ & $-$ & $4$ & $1,955$ & $4$ & $13,573$ & $-$ & $-$ & $-$ \\

EventCausality~\citep{do-etal-2011-minimally} & $25$ & $694$ & $17,326$ & $-$ & $746$ & $-$ & $-$ & $-$ & $-$ & $-$ & $1$ & $485$ & $-$ \\
BECauSE 2.0~\citep{dunietz-etal-2017-because} & $121$ & $4,038$ & $124,581$ & $4$ & $1,803$ & $109$ & $-$ & $-$ & $-$ & $-$ & $3$ & $110$ & $-$ \\
CaTeRS$^\ddag$~\citep{mostafazadeh-etal-2016-caters} & $320$ & $1,600$ & $-$ & $-$ & $2,708$ & $-$ & $-$ & $-$ & $-$ & $-$ & $3$ & $488$ & $-$ \\

EventStoryLine~\citep{caselli-vossen-2017-event} & $258$ & $4,316$ & $94,594$ & $7$ & $4,732$ & $-$ & $-$ & $-$ & $9$ & $8,111$ & $2$ & $4,584$ & $-$ \\
Causal-TB~\citep{mirza-etal-2014-annotating} & $183$ & $2,654$ & $63,811$ & $7$ & $6,811$ & $-$ & $-$ & $-$ & $13$ & $5,118$ & $1$ & $318$ & $-$ \\

Intelligence Community~\citep{hovy-etal-2013-events} & $100$ & $1,985$ & $51,093$ & $2$ & $3,919$ & $1,797$ & $-$ & $-$ & $-$ & $-$ & $-$ & $-$ & $4,586$ \\
HiEve~\citep{glavas-etal-2014-hieve} & $100$ & $1,354$ & $33,273$ & $6$ & $3,185$ & $2,734$ & $-$ & $-$ & $-$ & $-$ & $-$ & $-$ & $3,648$ \\
RED~\citep{ogorman2016richer} & $95$ & $2,719$ & $54,287$ & $-$ & $8,731$ & $2,049$ & $6$ & $893$ & $6$ & $4,209$ & $2$ & $1,147$ & $729$ \\
\ourdata & $4,480$ & $49,873$ & $1,275,644$ & $168$ & $112,276$ & $103,193$ & $4$ & $25,843$ & $6$ & $1,216,217$ & $2$ & $57,992$ & $15,841$ \\
     \bottomrule
\end{tabular}
}
\end{adjustbox}
\caption{Detailed statistics of \ourdata and existing widely-used datasets of all the ERE tasks. T-Link denotes temporal relations. C-Link denotes causal relations. $^\dag$: OntoNotes does not specify whether a mention is an entity or an event, so the \#Mention and \#Chain count both entities and events. $^*$: The majority of TempEval-3 is automatically annotated silver data. $^\ddag$: The original CaTeRS data is unavailable, so the statistics are taken from the original paper, and some statistics are missed.}
\label{tab:all_stat}
\end{table*}

\begin{table*}[!t]
\small
\centering
\begin{adjustbox}{max width=1\linewidth}
{
\begin{tabular}{l|lll|lll|lll}
\toprule
         & \multicolumn{3}{c|}{Train}                                                                & \multicolumn{3}{c|}{Development}                                                          & \multicolumn{3}{c}{Test}                                                                 \\ \cmidrule{2-10} 
         & \multicolumn{1}{c}{\#Doc.} & \multicolumn{1}{c}{\#Mention} & \multicolumn{1}{c|}{\#Chain} & \multicolumn{1}{c}{\#Doc.} & \multicolumn{1}{c}{\#Mention} & \multicolumn{1}{c|}{\#Chain} & \multicolumn{1}{c}{\#Doc.} & \multicolumn{1}{c}{\#Mention} & \multicolumn{1}{c}{\#Chain} \\ \midrule
ACE 2005 & $529$                      & $4,420$                       & $3,437$                      & $30$                       & $505$                         & $350$                        & $40$                       & $424$                         & $303$                       \\
TAC KBP  & $826$                      & $23,175$                      & $14,991$                     & $82$                       & $1,921$                       & $1,303$                      & $167$                      & $4,375$                       & $2,963$                     \\
\ourdata & $2,913$                    & $73,939$                      & $67,984$                     & $710$                      & $17,780$                      & $16,301$                     & $857$                      & $20,557$                      & $18,908$                    \\ \bottomrule
\end{tabular}
}
\end{adjustbox}
\caption{Data split statistics for datasets used in event coreference resolution experiments.}
\label{tab:split_coref}
\end{table*}

\begin{table*}[!t]
\small
\centering
\begin{adjustbox}{max width=1\linewidth}
{
\begin{tabular}{l|lll|lll|lll}
\toprule
         & \multicolumn{3}{c|}{Train}                                                                 & \multicolumn{3}{c|}{Development}                                                           & \multicolumn{3}{c}{Test}                                                                  \\ \cmidrule{2-10} 
         & \multicolumn{1}{c}{\#Doc.} & \multicolumn{1}{c}{\#Mention} & \multicolumn{1}{c|}{\#T-Link} & \multicolumn{1}{c}{\#Doc.} & \multicolumn{1}{c}{\#Mention} & \multicolumn{1}{c|}{\#T-Link} & \multicolumn{1}{c}{\#Doc.} & \multicolumn{1}{c}{\#Mention} & \multicolumn{1}{c}{\#T-LInk} \\ \midrule
TB-Dense & $22$                       & $1,212$                       & $7,553$                       & $5$                        & $150$                         & $898$                         & $9$                        & $350$                         & $2,299$                      \\
MATRES   & $182$                      & $6,684$                       & $6,332$                       & $73$                       & $4,431$                       & $6,404$                       & $20$                       & $746$                         & $837$                        \\
TCR      & $-$                        & $-$                           & $-$                           & $-$                        & $-$                           & $-$                           & $25$                       & $1,134$                       & $2,660$                      \\
\ourdata & $2,913$                    & $73,939$                      & $792,445$                     & $710$                      & $17,780$                      & $188,928$                     & $857$                      & $20,557$                      & $234,844$                    \\ \bottomrule
\end{tabular}
}
\end{adjustbox}
\caption{Data split statistics for datasets used in temporal relation extraction experiments.}
\label{tab:split_temporal}
\end{table*}

\begin{table*}[!t]
\small
\centering
\begin{adjustbox}{max width=1\linewidth}
{
\begin{tabular}{l|lll|lll|lll}
\toprule
         & \multicolumn{3}{c|}{Train}                                                                 & \multicolumn{3}{c|}{Development}                                                           & \multicolumn{3}{c}{Test}                                                                  \\ \cmidrule{2-10} 
         & \multicolumn{1}{c}{\#Doc.} & \multicolumn{1}{c}{\#Mention} & \multicolumn{1}{c|}{\#T-Link} & \multicolumn{1}{c}{\#Doc.} & \multicolumn{1}{c}{\#Mention} & \multicolumn{1}{c|}{\#T-Link} & \multicolumn{1}{c}{\#Doc.} & \multicolumn{1}{c}{\#Mention} & \multicolumn{1}{c}{\#T-LInk} \\ \midrule
\ourdata & $2,913$                    & $73,939$                      & $36,316$                     & $710$                      & $17,780$                      & $9,698$                     & $857$                      & $20,557$                      & $11,978$                    \\ \bottomrule
\end{tabular}
}
\end{adjustbox}
\caption{\ourdata split statistics for causal relation extraction experiments.}
\label{tab:split_causal}
\end{table*}

\begin{table*}[!t]
\small
\centering
\begin{adjustbox}{max width=1\linewidth}
{
\begin{tabular}{l|lll|lll|lll}
\toprule
         & \multicolumn{3}{c|}{Train}                                                                 & \multicolumn{3}{c|}{Development}                                                           & \multicolumn{3}{c}{Test}                                                                  \\ \cmidrule{2-10} 
         & \multicolumn{1}{c}{\#Doc.} & \multicolumn{1}{c}{\#Mention} & \multicolumn{1}{c|}{\#T-Link} & \multicolumn{1}{c}{\#Doc.} & \multicolumn{1}{c}{\#Mention} & \multicolumn{1}{c|}{\#T-Link} & \multicolumn{1}{c}{\#Doc.} & \multicolumn{1}{c}{\#Mention} & \multicolumn{1}{c}{\#T-LInk} \\ \midrule
HiEve    & $60$                       & $1,944$                       & $2,367$                       & $20$                       & $565$                         & $601$                         & $20$                       & $676$                         & $680$                        \\
\ourdata & $2,913$                    & $73,939$                      & $9,193$                    & $710$                      & $17,780$                      & $2,826$                    & $857$                      & $20,557$                      & $3,822$                    \\ \bottomrule
\end{tabular}
}
\end{adjustbox}
\caption{Data split statistics for datasets used in subevent relation extraction experiments.}
\label{tab:split_subevent}
\end{table*}

\section{Discussions on Genre Diversity}
\looseness=-1 \ourdata inherits all the documents of MAVEN~\citep{wang-etal-2020-maven}, which are all Wikipedia articles. One may wonder if \ourdata are diverse enough in genre and topic and if the ERE skills learned from the large-scale \ourdata can transfer to other ERE tasks (datasets). First, the original MAVEN work shows that the $4,480$ documents cover $90$ topics, such as Military conflict, Concert tour, etc. Hence we believe \ourdata also exhibits a good coverage for general-domain topics. Second, we conduct cross-dataset transfer experiments following~\citet{wang-etal-2020-maven}. By further fine-tuning the \RBT models previously trained on \ourdata, the (MUC) F1 scores increase $0.7$\%, $0.5$\%, $0.8$\%, $0.5$\%, $1.1$\%, $16.0$\%, $5.5$\%, $1.4$\% on ACE 2005, TAC KBP, TB-Dense, MATRES, TCR, Causal-TB, EventStoryLine, and HiEve, respectively. This shows that the general ERE skills learned from \ourdata are transferable and can help ERE on datasets in other genres, especially for these small-scale datasets. We encourage future works to explore the influence of genre gaps deeply.
\end{document}